\documentclass[runningheads]{llncs}
\usepackage{graphicx}
\usepackage{algorithm}
\usepackage{algpseudocode}
\usepackage{amsmath,amssymb} 

\usepackage{color}
\usepackage[width=122mm,left=12mm,paperwidth=146mm,height=193mm,top=12mm,paperheight=217mm]{geometry}
\usepackage{subcaption}
\usepackage{wrapfig,blindtext}
\usepackage{capt-of}
\begin{document}
\pagestyle{headings}
\mainmatter

\title{Lightweight Unsupervised Domain Adaptation by Convolutional  Filter Reconstruction}

\author{Rahaf Aljundi, Tinne Tuytelaars }
\institute{KU Leuven, ESAT-PSI - iMinds}

\maketitle

\begin{abstract}
End-to-end learning methods have achieved impressive results in many areas of computer vision. At the same time,  these methods still suffer from a degradation in performance when testing on new datasets that stem from a different distribution. This is known as the domain shift effect. Recently proposed adaptation methods   focus on retraining the network parameters. However, this requires access to all (labeled) source data, a large amount of (unlabeled) target data, and plenty of computational resources. In this work, we propose a lightweight alternative, that allows adapting to the target domain based on a limited number of target samples in a matter of minutes rather than hours, days or even weeks. To this end, we first analyze the output of each convolutional layer from a domain adaptation perspective. Surprisingly, we find that already at the very first layer, domain shift effects pop up. We then propose a new domain adaptation method, where first layer convolutional filters that are badly affected by the domain shift are reconstructed based on less affected ones. This improves the performance of the deep network on various benchmark datasets.
\end{abstract}

\section{Introduction}\label{sec:introduction}

In recent years, great advances have been realized towards image understanding in general and object recognition in particular, thanks to end-to-end learning of convolutional neural networks, seeking the optimal representation for the task at hand. 
Unfortunately, performance remarkably decreases when taking the trained algorithms and systems out of the lab and into the real world of practical applications. This is known in the literature as the domain shift problem: systems are typically deployed on new data that has different characteristics or has been gathered under different conditions than what was used for training. The default solution is to retrain or finetune the system using additional training data, mimicking as close as possible the conditions during testing. However, this brings an extra cost, first in terms of human effort to collect the new data and annotate it, next in terms of computational resources and expertise that need to be available to retrain the models. Moreover, it is not always feasible either, as conditions during test time may not be known well beforehand.

Overcoming this domain shift problem without additional annotated data is the main goal of unsupervised domain adaptation. It is the task of adapting a system trained on one data set (the {\em source} S) to be functional on a different data set (the {\em target} T). 
State-of-the-art methods for unsupervised domain adaptation of deep neural network architectures~\cite{gani2015domain,long2015learning} proceed by adding new layers to the deep network or learning a joint architecture in order to come up with representations that are more general and informative across the source and target domains. We will refer to these methods, that retrain the network using the information coming from the new (unlabeled) target samples, as {\em deep adaptation methods}. In this context (as in the context of finetuning), it has become common practice to retrain only the last layers of the deep network, supposing that the first layers are generic and not susceptible to any domain shift.

However, in spite of their good results on various benchmarks, these methods seem to be of limited value in a practical application. Indeed, deep adaptation methods require a lot of computation time, a lot of resources (powerful GPU-equipped servers), and a lot of unlabeled target data. This is in contrast to the typical domain adaptation setting, where we want networks trained on big datasets such as Imagenet to be readily usable by different users and in a variety of settings. For example, imagine a smart surveillance camera equipped with a pretrained recognition system that needs to be functional in a new context in spite of difficult lighting conditions, different backgrounds, etc. The camera does not have the resources on-board to retrain a deep convolutional network - it may not even have enough memory to store all the source data. Moreover, in many situations, we want the camera to be operational within a short time, so collecting a large set of target data samples is not an option either. The conditions to which we want to adapt, may also vary over time (e.g. lighting conditions in an outdoor application), so if the adaptation process takes too long, the new model may already be outdated by the time it becomes available.

So instead, we advocate the need for {\em light-weight domain adaptation schemes}, that require only a small number of target samples and can be applied quickly without heavy requirements on available resources, in an {\em on-the-fly} spirit. Using only a few samples, such a system could adapt to new conditions at regular time intervals, making sure the models are well adapted to the current conditions.
The simpler sub-space based domain adaptation methods developed earlier for shallow architectures \cite{fernando2013unsupervised,gong2013connecting,gopalan2011domain} seem good candidates for this setting. Unfortunately, when applied to the last fully connected layer of a standard convolutional neural network, they yield minimal improvement~\cite{tommasi2015deeper,donahue2013decaf}. 
In this work, we start by analyzing the different layers of a deep network from a domain adaptation perspective (section~\ref{sec:Analysis}). First, we show that {\em domain shift does not only affect the last layers of the network}, but can already manifest itself as early as the very first layer. Second, we show that the {\em filters 
exhibit different behavior in terms of domain shift}: while some filters result in a largely domain invariant representation, others lead to very different results for the source and target data. Based on this analysis,  we propose a new light-weight domain adaptation method, focusing just on the first layer of the network (section~\ref{sec:FilterReconstruction}). For a given target data sample, it aims at {\em reconstructing the output of the filters affected by domain shift} such that their new output is more similar to the response given under the training conditions (i.e., the source dataset).  We evaluate the method on various benchmarks: the Office dataset~\cite{saenko2010adapting}, Mnist~\cite{lecun1998gradient}, a Photo-Art dataset  and the German traffic sign dataset GTSRB~\cite{sermanet2011traffic} (section~\ref{sec:Exp}). The proposed method can adapt the learned network and improve the raw performance, even when only a few (unlabeled) samples from the new domain are available. It takes minimal time 
and does not involve  parameter tuning. But first, let us describe related work (section~\ref{sec:related}).

\section{Related Work}
\label{sec:related}
{\bf Shallow DA \ \ } So far, domain adaptation (DA) has mostly been studied in the context of image representations based on handcrafted features. Methods (see~\cite{patel2015visual} for a survey) tackle the problem in different ways, such as the feature augmentation scheme of~\cite{daume2009frustratingly} or instance reweighting~\cite{chu2013selective,huang2006correcting}, that tries to correct the shift by re-weighting the source samples based on their similarity with the target domain. Another interesting line of work is the use of a latent feature space~\cite{pan2011domain,long2013transfer}, that has led to the development of subspace-based DA methods~\cite{fernando2013unsupervised,gopalan2011domain,aljundi2015landmarks,sun2015subspace,hoffman2014continuous}. 
Especially the work of~\cite{hoffman2014continuous} 
is worth mentioning here, as it aims at adapting a model in an online fashion, somewhat similar in spirit to our work.
Most of these methods have mainly been evaluated on the Office benchmark~\cite{saenko2010adapting}  that comes with precomputed SURF features. 
However,
when applied to deep features (i.e., activations of the last layer of a pretrained convolutional neural network), that capture more high-level object information rather than edges and gradients, they do not seem as powerful as before~\cite{tommasi2015deeper,donahue2013decaf}. 
Therefore, more recent works use deep learning like methods to reduce the domain shift. 


\noindent
{\bf Deep DA \ \ } Deep adaptation methods try to integrate the adaptation within the learning process: \cite{chopra2013dlid} learns a  joint architecture between the source and target data, while in \cite{ghifary2014domain}, a denoising auto-encoder is used to learn from unlabeled target data after which a two layer network with minimum mean discrepancy (MMD) as an adaptation loss is trained. 
The performance of these two methods is limited, however, due to the use of relatively shallow architectures. They can easily be outperformed by finetuning a deeper network on the source data. 
Tzeng {\em et al.}~\cite{tzeng2014deep} proposed to add a fork to the deep network that first determines which layer to use and then decides the dimension of that layer based on a combination of the classification loss on the source data and domain confusion loss, again using  MMD. This method needs to fine-tune different networks for choosing the best dimensionality and the adaptation process is limited to the choice of the layer in which the joint loss is minimized. A more comprehensive  deep learning based approach~\cite{long2015learning} suggests to freeze the first layers, as it supposes that these layers are global and are not prone to the domain shift. Then, to retrain the higher layers, a multi-layer adaptation regularizer based on multi-kernel MMD is added to improve the transferability of the features. 
In contrast, we show that  the first layers are not  immune to domain shifts even though they are   generic feature extractors 
and that, by improving only the first layer features, we can already improve the performance of the whole network.
Finally, \cite{gani2015domain} shows a significant improvement through the use of a deep network along with a domain regressor. This is done by adding a sub-network consisting of parallel layers to the classification layer. These layers use a different loss that minimizes the discrimination ability  between source and target instances, which is incorporated in the back propagation learning scheme.\\
Even though these latter, deep-learning based approaches give good results and go along with the current research line (deep learning), it is quite challenging to apply them directly. First, one needs to determine the number of layers that need to be added for the adaptation process, which is difficult without access to the labels. Second, even if we suppose that the number of layers and all the other parameters are given, these methods require to retrain the whole network, which may involve millions of parameters and training samples. This retraining could take a week, so it is impossible  to apply the model in a reasonable time. 

In conclusion, we see that most of the methods try to relax the problem, either with some assumptions about the characteristics of the used features or by having access to different parameters and a lot of time and computational resources. 
In contrast, we propose a simple yet efficient method that: \textbf{1)} attempts  to reduce the shift and improve the network performance {\em without retraining} the network or any other classifier, and \textbf{2)} needs only a few (unlabeled) samples from the target domain to be functional. We do so by adapting the first layer of the network - which goes against  common belief and practice. Below, we explain our findings w.r.t. the different layers and then proceed to  the proposed method.

\section{Analysis of domain shift in the context of deep learning}\label{sec:Analysis}
 As mentioned above, deep adaptation methods typically assume that the first layers are generic and need no adaptation, while the last layers are more specific to the dataset used for training and thus sensitive to the  shift between the two domains. Therefore, most of the new adaptation methods tend to adapt only the last layers and freeze the first layers.
This assumption is based on the fact that the first layers mainly detect colors, edges, textures, etc. - features that are generic to all domains. On the other hand, in the last layers we can see high level information about the objects, which might be domain-specific. 
However, {\em given that the feature extraction method is generic, does that mean the features are not conveying any domain shift between the different datasets?} To answer this question, we perform a thorough analysis of the output of each layer. For this purpose, we use Alexnet~\cite{krizhevsky2012imagenet} pretrained on Imagenet. This network shows remarkable performance and has been trained to recognize a thousand objects. 

Domain shift can be caused by different factors which can be mainly divided into: 1. a {\em generic shift}, due to having objects with different appearance or captured from different view points (as one can expect when using different datasets for training and testing), and 2. a {\em low-level shift}, uniformly affecting the images in the dataset due to e.g. lighting conditions, colors, etc. To study these two types of shift, we use the Office dataset~\cite{saenko2010adapting} which is the de facto standard benchmark for domain adaptation. 
First,  we use the sets Amazon and Webcam (where Amazon contains images gathered from the Amazon website and Webcam contains images captured by a web camera). With this setting, we cover the generic shift (in this case mostly due to the white background on Amazon compared to non-uniform background in the Webcam data set, as well as different objects with different appearance). In addition, to study the low level shift, we created a gray scale set that contains the same images as in the Amazon dataset but converted to gray scale. We call it Amazon-Gray. We consider two adaptation cases: Amazon $\rightarrow$ Amazon-Gray and Amazon $\rightarrow$ Webcam.  
We start by fine-tuning AlexNet on the Amazon dataset (the source). Then, we consider each convolutional layer as an independent feature extraction step. Each layer is composed of a set of filter maps. We consider one filter from one layer at a time, and consider, for this analysis, the corresponding filter map as our feature space. Each instance (image) is a point in that space. For example, the first convolutional layer is composed of 96 filter maps, each of size [55x55]. We reshape each [55x55] filter map into a feature vector of length 3025 and consider this the feature representation of the image. Now we want to find out whether source and target samples follow the same distribution in this feature space, or not.\\

\begin{figure}[t]
    \centering
    \begin{subfigure}[b]{0.3\textwidth}
        \includegraphics[width=\textwidth]{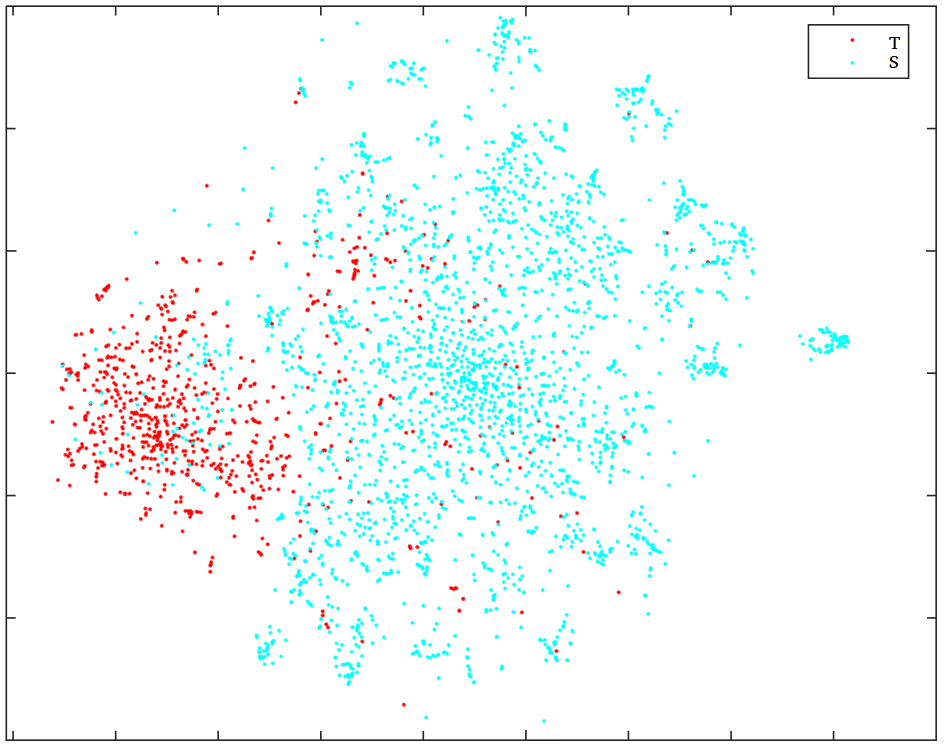}
      
    \end{subfigure}
    ~ 
    \begin{subfigure}[b]{0.3\textwidth}
        \includegraphics[width=\textwidth]{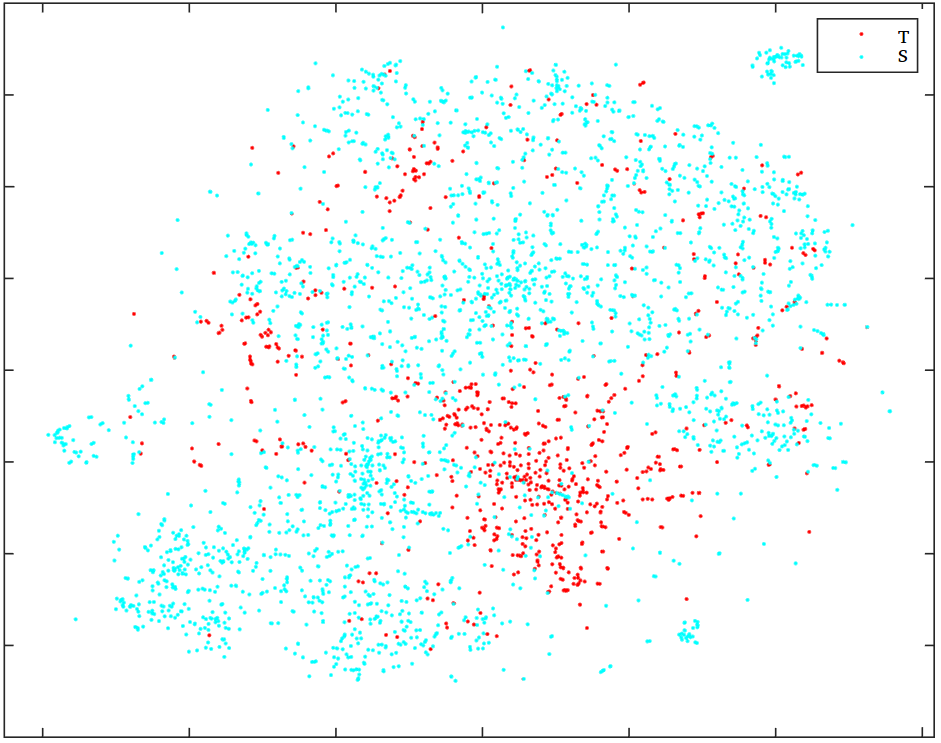}
      
    \end{subfigure}
    ~ 
    \begin{subfigure}[b]{0.3\textwidth}
        \includegraphics[width=\textwidth]{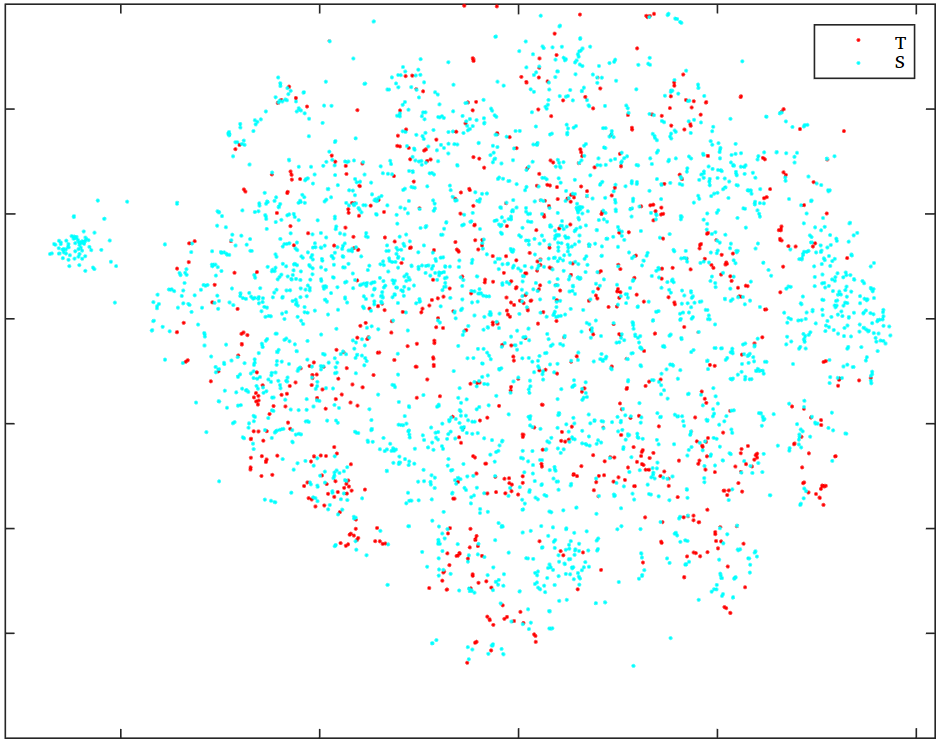}

    \end{subfigure}
    \caption{From left to right: visualization of source points (blue) and target points (red) for three different filters from conv1, conv2, and conv3 resp., for the Amazon$\rightarrow$Webcam setup.}\label{fig:webcam}
    \vspace*{-0.3cm}
\end{figure}

\begin{figure}[t]
    \centering
    \begin{subfigure}[b]{0.3\textwidth}
        \includegraphics[width=\textwidth]{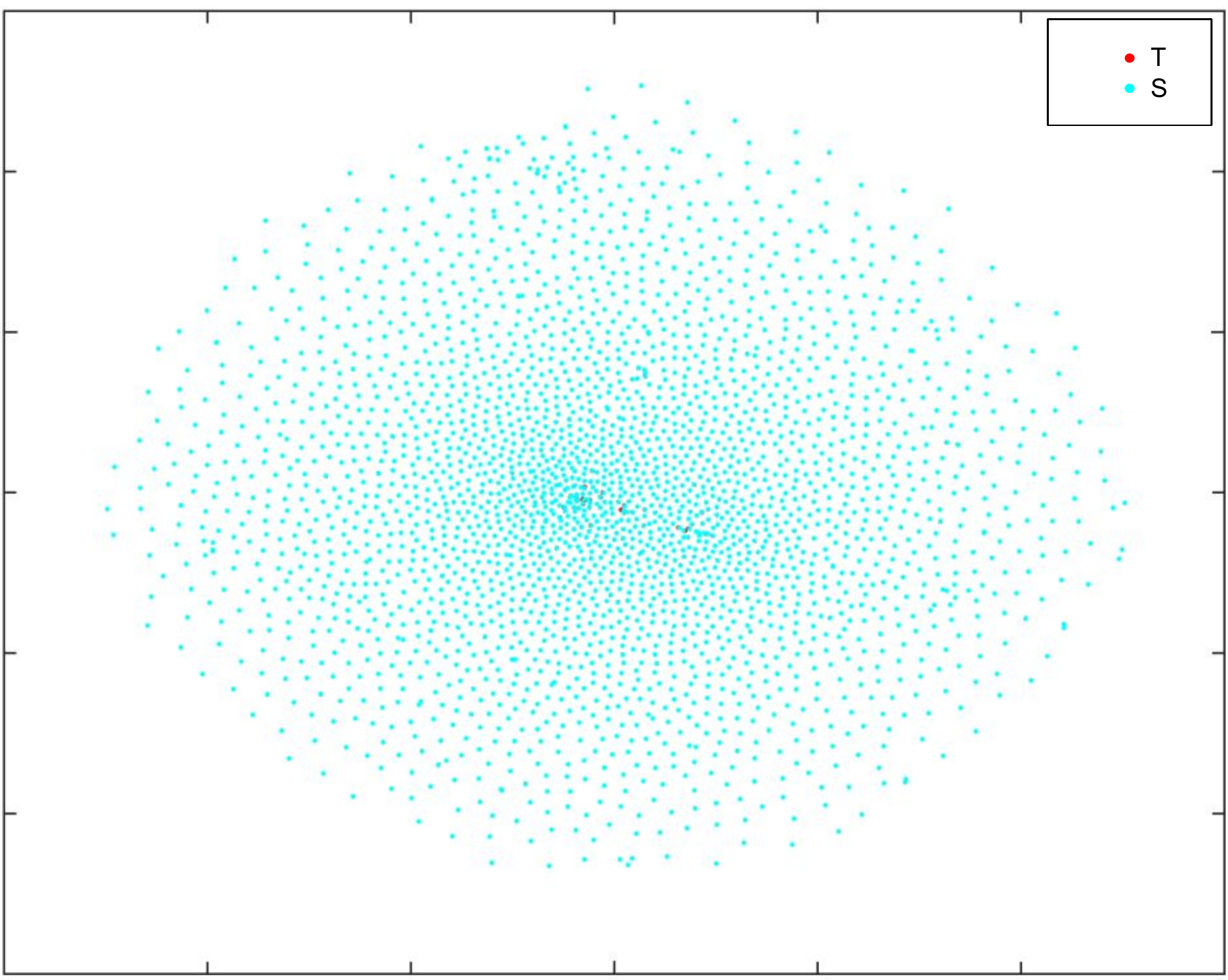}
      
    \end{subfigure}
    ~ 
    \begin{subfigure}[b]{0.3\textwidth}
        \includegraphics[width=\textwidth]{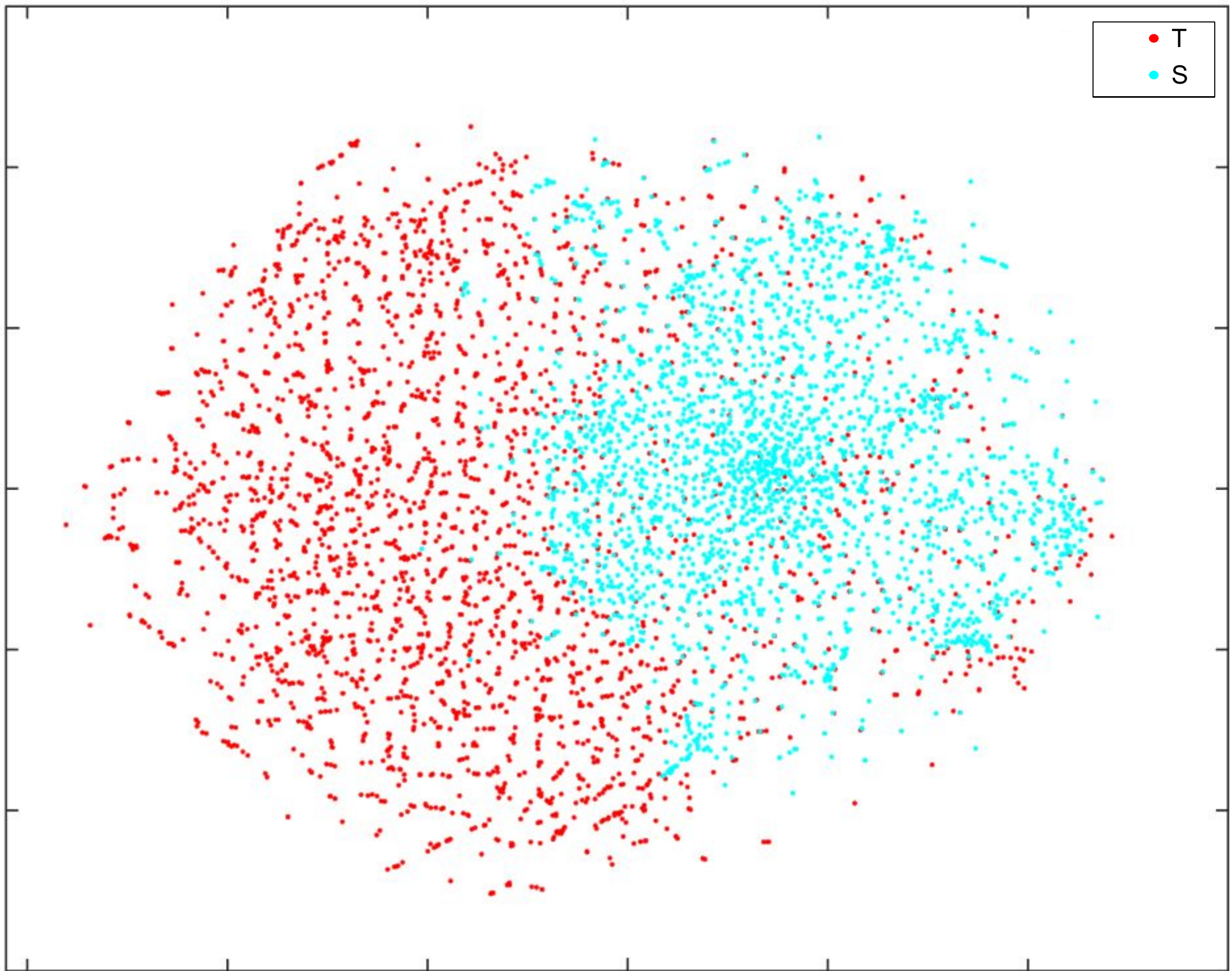}
      
    \end{subfigure}
 
    \caption{On the left, a good conv1 filter with samples nicely matched from the two domains Amazon and Amazon-Gray. On the right, a conv1 filter affected by the domain shift with different distributions for the two domains. }\label{fig:gray_bad_good}
    \vspace*{-0.3cm}
\end{figure}

\noindent
{\bf Qualitative analysis \ \  } As a first step, we visualize the points in each feature space. To do so, we use the tSne package \cite{van2008visualizing} that is mostly suitable for visualizing high dimensional data. We visualize the points for each domain adaptation couple.
Figure~\ref{fig:webcam} shows the visualizations of some example filters from different convolutional layers for Amazon$\rightarrow$Webcam. Figure~\ref{fig:gray_bad_good} shows the visualization for two different filters of the first layer for Amazon$\rightarrow$Amazon-Gray. Please refer to the supplementary materials for more visualizations. 
The behavior of the filters is quite distinct. In general, the higher the layer, the more overlap between the source and target features we can observe.\\
 
\begin{figure}[t]
\centering
\includegraphics[width=0.7\textwidth]{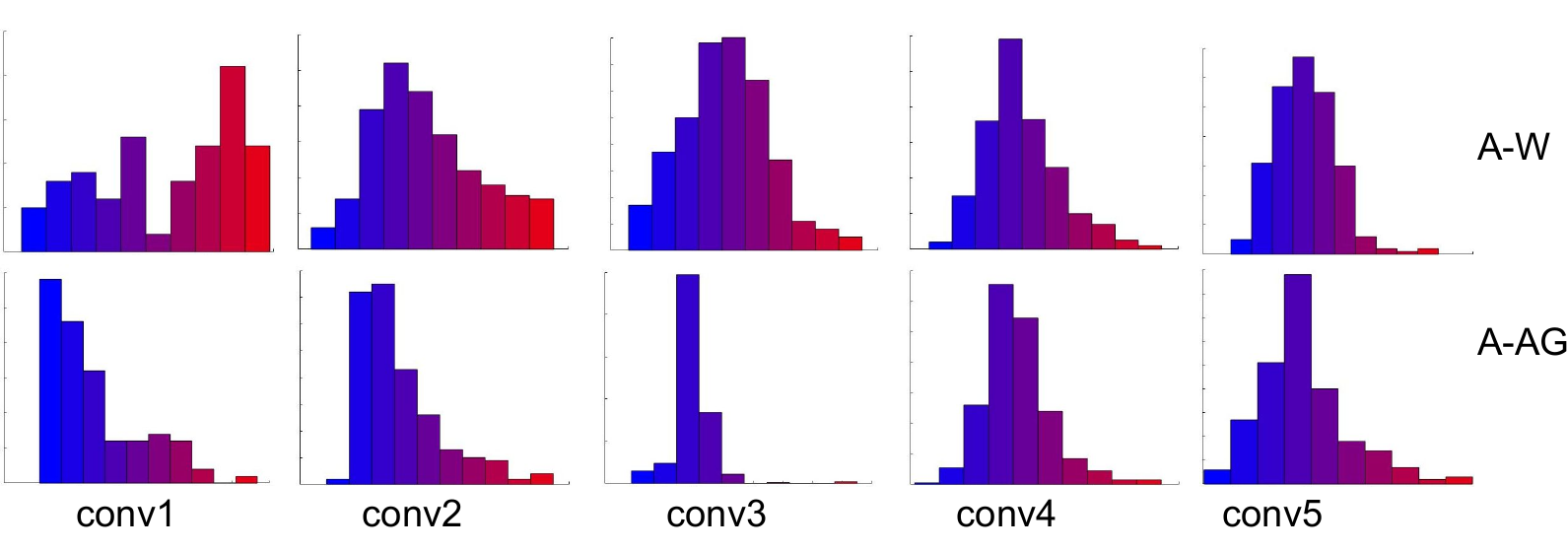}
  \caption{The histograms of the filters' H-divergences for different layers.}
   \vspace*{-0.5cm} 
  \label{fig:hdiv}
\end{figure}

\noindent
{\bf H-divergence \ \ }
In addition to visualizing the output of the filters in each layer of the deep network, we study the H-divergence w.r.t. each layer $\slash$ filter as well. H-divergence was first introduced by Ben David {\em et al.}~\cite{ben2007analysis} as a measure of the discrepancy between the source distribution $D_S$ and target distribution $D_T$. The H-divergence can be written as follows:
\begin{equation}
d_{H \Delta H}(D_S,D_T)=\underset {h\in H} {2sup}|Pr_{x\sim D_S} (h(x)=1)-Pr_{x\sim D_T}(h(x)=1) |
\end{equation}
where $h \in H$ is a characteristic function that is learned to discriminate between samples generated from the source distribution and those generated from the target distribution. Clearly, the bigger the error, the bigger the divergence.
In our case, we use an SVM classifier with linear kernel and a fixed C value for all the different computations of the H-divergence. We use a similar feature representation as for the visualization method above. 
We repeat the process for all filters in all layers. In Figure~\ref{fig:hdiv}, we show the histograms of the filters' H-divergences w.r.t. each layer regarding the two study cases Amazon$\rightarrow$Webcam and Amazon$\rightarrow$Amazon-Gray. We encode the value of the H-divergence by the color where blue indicates a low H-divergence (= "good" filters) while red indicates a high H-divergence (= "bad" filters).\\



\noindent
{\bf Discussion \ \ }
From Figures~\ref{fig:webcam} and~\ref{fig:hdiv}, we can conclude that, in contrast to common belief, the {\em first layers are susceptible to domain shift even more than the later layers} 
(i.e., the distributions of the source and target filter outputs show bigger differences in feature space, resulting in larger H-divergence scores).
%
%
%
Indeed, the filters of the first layers are similar to HOG, SURF or SIFT (edge detectors, color detectors, texture, etc.); they are generic w.r.t. different datasets, i.e. they give representative information regardless of the dataset. However, this information also conveys the specific characteristics of the dataset and thus the dataset bias. As a result, when the rest of the network processes this output, it will be affected by the shown  bias, causing a degradation in performance.

Especially in the first layer of the convolutional neural network, we see large differences between different filters (Figures~\ref{fig:gray_bad_good} and~\ref{fig:hdiv}). For some filters, the samples are almost perfectly matching while others have very different distributions. We do not observe such close-to-perfect matchings in the later layers, as their input is affected by the domain shift in some of the first layer filters. Based on this analysis of the domain bias over different layers, we believe that {\em a good solution of the domain adaptation problem should start from the first layers} in order to correct each shift at the right level rather than waiting till the last layer and then try to match the two feature spaces.\\

\noindent
{\bf Our DA strategy \ \ }
Based on these findings and keeping in mind our goal of having a lightweight method that compensates for the domain shift without the need to retrain the network or any other heavy computations, we suggest the following strategy:
\begin{itemize}
\item compute the divergence of the two datasets with respect to each filter  as a measure for how “good” each filter is, and
%
%
\item  use the “good” filters to reconstruct the output of the “bad” filters.
\end{itemize}

\begin{wrapfigure}{r}{0.4\textwidth}
   \vspace*{-0.9cm} 
    \includegraphics[width=0.4\textwidth]{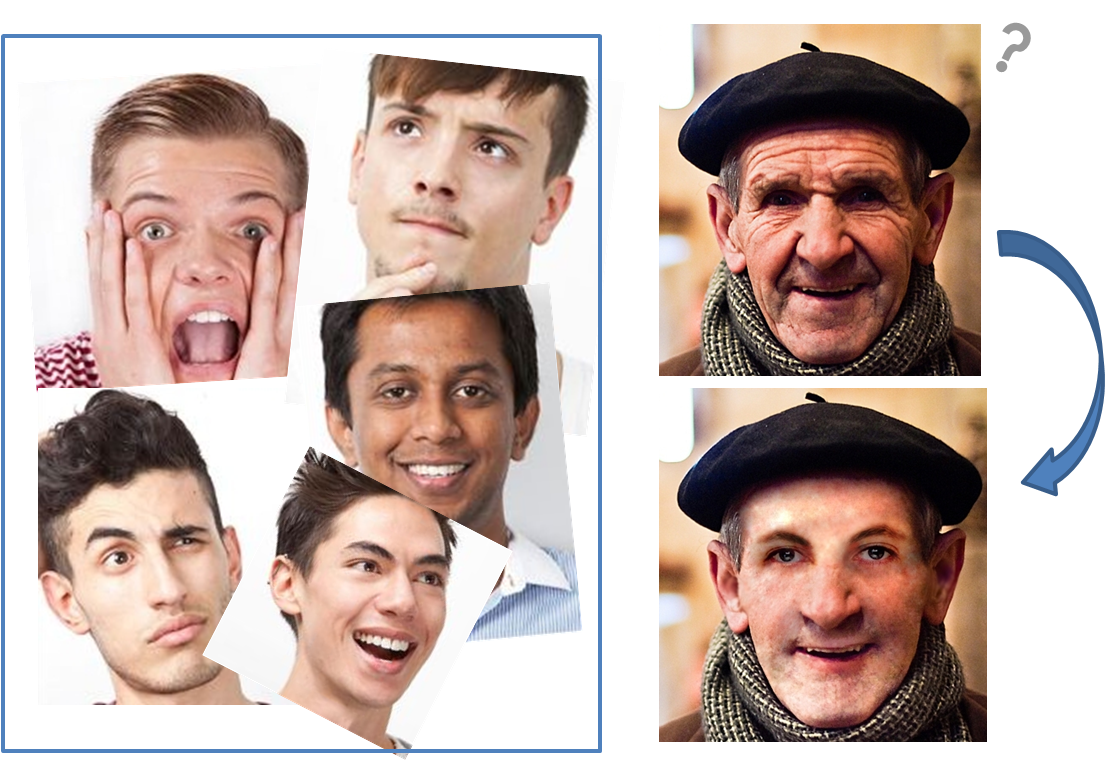}

  \caption{The intuition behind filter reconstruction }
   \vspace*{-1cm} 
  \label{fig:idea}
\end{wrapfigure}

By reconstructing the bad filters output, we mean: starting from a target image as input, for each bad filter, re-estimate its response map such that it becomes more similar to the response map of a source image for the same filter. To achieve such reconstruction, we rely on the target filter maps of the good filters, for which we know the responses from the two datasets to be similar. Figure \ref{fig:idea} illustrates the concept behind the filter reconstruction. Suppose a system is trained to recognize facial expressions based on a set of young faces (i.e., the source) but now has to be applied on elder people's faces (i.e., the target). Because of the wrinkles in the old faces, the recognition will be inaccurate (i.e., there's a domain shift problem). Some of the low level filters will not be affected much by the domain shift (e.g. color filters), while others will show large domain shifts (e.g. texture filters). Now we propose to use information in common with the young faces (color) to reconstruct the filter maps of the bad filters (texture). This corresponds to removing the wrinkles, which in turn allows a better recognition. In the following, we will explain the filter reconstruction scheme and then test the proposed method on a variety of datasets. For simplicity, we just focus on the first layer from now on. 

\section{Filter Reconstruction}\label{sec:FilterReconstruction}
We start from a set of filters from a given layer. Some of these filters are more prone to  domain shift than others.  We want to determine the bad filters in order to reconstruct their output given the good filter responses. Here "good" and "bad" are from the perspective of the adaptation problem in hand.  We aim at designing an optimization problem that simultaneously determines the bad filters and identifies a set of filters that can be used for reconstructing the bad filters output. In order to achieve this, we consider one filter at a time. The filter under study is considered either good (and, hence, retained) or bad (in which case better filters are selected to reconstruct its output). \\

\noindent
{\bf A feature selection analogy \ \ }
To explain our proposed solution, we use the analogy with a feature selection operation where we are given a set of features$\slash$ predictors (all filter maps) and a desired output $\slash$ response (the filter map under scrutiny).  We want to select the set of features based on which we can predict the given response. Clearly, if we have the response itself as a feature in the features set, it will be directly selected as it is the most correlated with the output (i.e. itself). But now we add another criterion to the feature selection problem, that indicates how good a feature is regarding the new problem in which the predicted output will be used.  This new problem is classifying new samples from the target dataset. The additional selection criterion is based on the resulting shift with respect to the filter in hand, i.e the divergence shown by this filter. First, we describe our divergence measure, then use it in the selection process.  \\

\noindent
{\bf KL-divergence \ \ }
We need a measure that can be computed efficiently and that can give us an indication on how good or bad the filter is in terms of the adaptation problem. For that purpose,  we  estimate the probability distribution of the filter response given source data as input, $P_S$, and likewise for the target data (or a subset thereof), $P_T$.  We then use the KL-divergence~\cite{kullback1951}~\cite{kullback1987letter}.  That is a measure of the difference between two probability distributions, in our case $P_S$ and $P_T$.  
It estimates the amount of information lost when using the source probability distribution to encode the target probability distribution.
For the case of discrete probability distributions, KL-divergence is computed as follows:
\begin{equation}
D_{KL}(P_T\parallel P_S)=\sum_i P_T(i)log \frac{P_T(i)}{P_S(i)}
\end{equation}
In the context of our problem,  for each filter, we estimate the distribution of the source samples and the distribution of the target samples for that filter response and then compute the KL divergence between the two probability distributions. This gives us a KL divergence value associated with each filter. We use this value as our additional criterion in the filter selection operation.\\
\noindent
{\bf Filter Selection \ \ } We  want to  select the set of features (filters) that are going to be used in a regression function that predicts the output of the filter in hand.  
Here, we do not consider the entire filter map, but rather the filter response at each point of the filter map separately, where, given the response of the other filters at this point we want to predict the current response. We use the source data as our training set where our aim is to reconstruct the source like response of a bad filter given the output of the good filters.
Going back to the literature, feature selection for regression has been studied widely. Lasso~\cite{tibshirani1996regression}  and Elastic net~\cite{zou2005regularization} have shown good performance. The two methods differ in their regularization strategy; while Lasso introduces the L1 norm regularization that will ensure the sparsity of the selected set of features, the Elastic net adds another  L2 norm regularization term. By doing so, the Elastic net overcomes  the case when  the number of features is bigger than the number of samples and encourages grouping of features as well. In our case, we always have a set of source samples bigger than the number of filters, and we don't want to group the selected filters. Therefore, we  favor Lasso as it introduces the sparsity which is essential in our case to select as few and effective filters as possible.
Having  the response $y$ and the set of predictors $x$, the main equation of Lasso can be written as follows:
\begin{equation}
  B^*=argmin_B \{ \sum _{i=1}^{n} (y_i -\beta_0  -\sum_{j=1}^{p}x_{ij}\beta_j)^2  + \lambda\sum_{j=1}^p |\beta_j| \}
  \end{equation}
where $\beta_0$ is the residual, $B = \{\beta_j\}$ the estimated coefficients, $n$ the number of source samples, $p$ the number of filters,  and $\lambda$ a tuning parameter to control the amount of shrinkage needed. The bigger  the value of $\lambda$,  the more we steer the coefficients $\beta_j$ towards zero values.
What we need to do next is to insert our additional selection criterion, i.e. the KL-divergence, where for each filter $x_j$, we have computed a KL divergence value, $\Delta_j^{KL}$. We will use this divergence value to guide the selection procedure. This can be achieved by simply plugging the $\Delta^{KL}_j$ value in the $L_1$ norm regularization as follows: 
  \begin{equation}
  B^*=argmin_B \{ \sum _{i=1}^{n} (y_i -\beta_0  -\sum_{j=1}^{p}x_{ij}\beta_j)^2  + \lambda\sum_{j=1}^p |\Delta_j^{KL} \cdot{ \beta_j|} \}
  \end{equation}
Solving this optimization problem, we obtain the weights vector $B^*$, with a weight $\beta^*_j$  for each filter, including the filter we try to reconstruct itself. If the filter in hand has a non-zero weight, that means it is considered  a good filter and we will keep its value. On the other hand, if the filter has zero weight then it will be marked for reconstruction and the filters with non-zero weights are used for this purpose.
The above optimization problem can be nicely solved using coordinate descent where we update the gradient  w.r.t.  one filter at a time. In the normal Lasso setup~\cite{friedman2010regularization}, the regularization paths are used to select lambda. In our case, we have another term, i.e. the divergence, associated with each selection. We choose the lambda that gives us an optimal combination of divergence between the source points and the reconstructed target points at one hand  
and the error of the reconstructed source  on the other hand.    The coordinate descent optimization is quite fast and scalable as it doesn't need to keep the training data in memory.\\
  
\noindent  
{\bf Reconstruction \ \ } The Lasso optimization process is very efficient in selecting the features, but the weights that are estimated by the method are not the most accurate for prediction as they are controlled by  $\lambda$ . A common practice is to use Lasso for the variable selection step and then use another  regression method for the prediction. Therefore, after selecting the set of filters to be used for reconstruction, we use the linear regression method to predict the filter output $y_{l}$ given the responses of the selected filters. 
The linear regression is in its turn simple and efficient to compute.
As a result, we obtain the final set of coefficients $B_{l}$ 
for each bad filter $y_l$. Algorithm \ref{algo} summarizes the filter selection and reconstruction procedure. \\

\noindent
{\bf Prediction \ \ } At test time we receive a target sample $x_t$. We pass it through the first layer and obtain the response of each filter map. Then, for each bad filter $y_l$, we use the responses of its own  selected set of filters to predict a source like response given the coefficients $B_{l}$. After that, we replace each bad filter value (a point in the response of the bad filter map) by the predicted response and pass the reconstructed data to the next layer up to prediction.
\begin{algorithm}

  \caption{Filter Reconstruction}\label{algo}
  \begin{algorithmic}[1]
  \State Input $S,T$\Comment{The first layer output of the source and of the given target samples}
      \State $\Delta^{KL}\gets$  compute-KL$(S,T)$
      \For{\texttt{$l$ in Filters }}\Comment{Loop around the filters of the first layer }
        \State $B^*=Lasso(S,\Delta^{KL},l)$
        \If {$B^*(l)=0 $}
    		\State $BadFilters(l)\gets 1$ \Comment{Add it to the set of bad filters}
            \State $Selected(l)\gets B^*\neq 0$  \Comment{The selected set of filters for its reconstruction}
            \State $B_{l} \gets LinearRegression(S(l),Selected(l))$ 
            \State $PredB(l)\gets B_{l}$ \Comment{The set of coefficients for all the bad filters}
		\EndIf
      \EndFor
      \State \textbf{return} $BadFilters, Selected, PredB$ 
 
  \end{algorithmic}

\end{algorithm}
  \vspace*{-1cm} 
\section{Experiments}\label{sec:Exp}
In this section, we design different experiments in order to see the complete picture of the filter reconstruction method and to have a clear understanding of when and where it is a good practice to use.  We test the performance of the method on different datasets and compare  with multiple baselines.

\subsection{Setup $\&$ Datasets}
\textbf{Office benchmark} \cite{saenko2010adapting}: it contains three sets of samples: i) Webcam, composed of images taken by a webcamera,  ii) DSLR, containing images captured by a digital SLR camera, and iii) Amazon which contains images of office related objects downloaded from the Amazon website. In addition, we will use the gray scale version of the Amazon data.  The main task is object recognition. For the first set of experiments, we will use Alexnet~\cite{krizhevsky2012imagenet} pretrained on Imagenet and fine-tuned on the Amazon data. We will be dealing with three adaptation problems: Amazon$\rightarrow$ Amazon-Gray,  Amazon $\rightarrow$ Webcam and Amazon $\rightarrow$ DSLR. We do not consider DSLR$\leftrightarrow$Webcam, as with deep features the shift between the two sets is minimal.
\begin{wrapfigure}{r}{0.4\textwidth}
  \vspace*{0.3cm}
    \includegraphics[width=0.4\textwidth,scale=0.5]{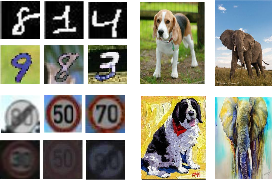}
  \caption{Examples of the Mnist-MinstM, Syn-Dark and Photo-Art adaptation couples}
    \label{fig:examples}
   \vspace*{-0.8cm}
\end{wrapfigure}
%

\noindent
\textbf{Mnist$\rightarrow$ MnistM}: here, we train a network on Mnist dataset for hand written digits recognition that are black \& white, while the target test is the MnistM dataset, composed of the same digits as Mnist~\cite{lecun1998gradient} but blended with patches from real image~\cite{arbelaez2011contour}. We followed the procedure described by~\cite{gani2015domain} and further verified by having the same results. \\
\textbf{Synthetic traffic signs$\rightarrow$ Dark illumination}:  in this setting, we want to imitate the real life condition in which we train a system on a large dataset of synthetic images that could as much as possible mimic all the different conditions, and then test it on a dataset affected by a domain shift. The task here is traffic sign recognition where we train on a synthetic dataset of traffic signs \cite{moiseev2013evaluation} composed of 100.000 samples and test on a subset of the German traffic sign dataset \cite{sermanet2011traffic} that has been captured under very dark light conditions (see Figure~\ref{fig:examples}). We follow a single column traffic sign network architecture similar to \cite{cirecsan2012multi}. \\
\textbf{Photo$\rightarrow$Art}: finally we examine a different shift, which is the case of having a network trained on photos of real objects and then test it on paintings. A similar idea has been  introduced in \cite{cai2015beyond} in the context of cross depiction problem where the introduced target set was gathered from different domains, i.e. clip art, painting, cartoon and sketches.
Here we fine-tune Alexnet on a set of 7 classes of animals, whose real photos we downloaded from the internet  using a commercial image search engine. The used categories are: dog, elephant, zebra, lion, cow, deer and horse.
In a similar way, we gathered paintings of the same categories from the internet.The training and test sets have around 40 images per category for each set and  will be available online. We apply our domain adaptation method to the first layer of the fine-tuned network.
See figure \ref{fig:examples} for an example of the different adaptation problems. For the previously described adaptation problems, we used 10$\%$ of the target dataset as our available target samples.
\paragraph{\textbf{Baselines:}} We compare with the following baselines: \textbf{No adaptation(NA)} by testing the network fine-tuned on the source dataset directly on the target set without adaptation.\textbf{DDC} method \cite{tzeng2014deep} that only adapts the last layer by selecting the feature space dimension that shows the MMD between the source and target domains.\textbf{ Subspace Alignment(SA):} unsupervised subspace alignment\cite{fernando2013unsupervised} is a simple method, yet shows good performance. To make a fair comparison, we take the activations of the last fully connected layer before the classification layer and use them as features for this baseline. We perform the subspace alignment and retrain an SVM with a linear kernel on the aligned source data, then use the learned classifier to predict the labels of the target data. We tried all the different dimensions between 20 and 60 and report the best result regarding the target classification task. For completeness, we also show the result of the SVM classifier trained on the source features before alignment(\textbf{SVM-fc7}).
\textbf{ SA - First Convolutional:} we adapt the subspace alignment method so it can be applied to the  first layer of the convolutional neural network. To avoid having to retrain a classifier, we adapt the target data to the source data (instead of vice versa), w.r.t. the activations of the first convolutional neural network. 
In the original subspace alignment problem, the source subspace is aligned with the target subspace using the mapping matrix $M=X_S^TX_T$  where $X_S $ and $X_T$ are the biggest d principal components of the source and the target. Source and target samples can then be compared using $y_S  X_s X_s'  X_T X_T' y_T'$. Here, we therefore replace a filter response vector $y_T$ by $y_T X_T X^T X_S X_S^T$. 
%
Again, we tried different dimensions for the subspaces, between 20 and the size of the filter maps, i.e. 96 in Alexnet.

\subsection{Results and discussion}
Table \ref{tab1} shows the results of the experiments on the  Office dataset.  In spite of the method's simplicity  and the fact that it is just active on the first layer, we systematically improve over the raw performance obtained without domain adaptation - especially in the case of Amazon-Gray where it is  a low level shift and the method  adapts by anticipating the color information of the target dataset, i.e. reconstructing the color filters. In the case of Amazon-Webcam and Amazon-DSLR, the method tries to ignore the background of Webcam and DSLR datasets that is different  from the white background in Amazon dataset. 

\begin{table}
 \vspace*{-0.8cm} 
\caption{\label{tab1}Recognition accuracies on Office dataset}
\centering
\begin{tabular}{ | c | c| c | c |}
\hline
Method & Amazon$\rightarrow$Webcam& Amazon$\rightarrow$DSLR & Amazon$\rightarrow$Amazon-Gray \\ \hline
CNN(NA) & 60.5 & 65.8& 94.8  \\ \hline 
DDC\cite{tzeng2014deep} &61.8&64.4& -\\ \hline
SVM-fc7(NA) & 60.5 & 61.5&95.0  \\ \hline 
SA\cite{fernando2013unsupervised} & 61.8 & 61.5 &95.2  \\ \hline 
SA(First Convolutional)&61.5&65.8& 95.1\\ \hline 

Filter Reconstruction(Our) &  \textbf{62.0} & \textbf{ 67.2} &  \textbf{ 97.0}  \\ \hline 

\end{tabular}
 \vspace*{-1.5cm} 
\end{table}

\begin{figure}
   \vspace*{-0.3cm} 
    \centering
 \includegraphics[width=0.8\textwidth,scale=0.5]{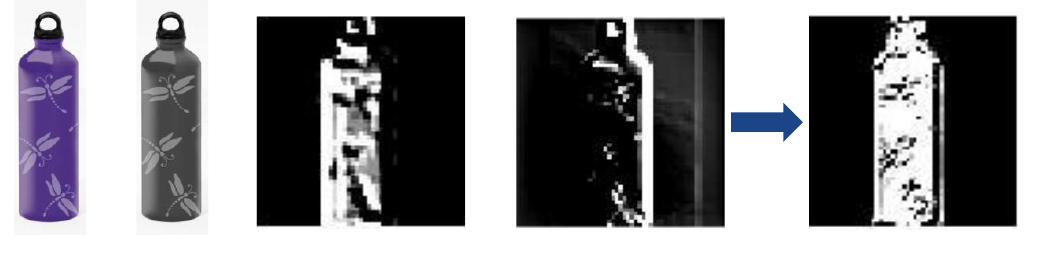}
 \caption{From left to right: a sample image from Amazon, the corresponding gray image, the output of a bad filter w.r.t. the color image, the output of the same bad filter w.r.t. the gray image and the reconstructed output.}\label{fig:gray_reconstruction}
  \vspace*{-1.2cm} 
\end{figure}

\begin{figure}
 
    \centering
    \begin{subfigure}[b]{0.15\textwidth}
        \includegraphics[width=\textwidth]{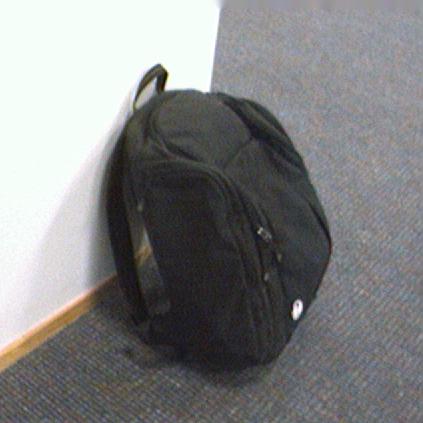}
      
    \end{subfigure}
    ~ 
    \begin{subfigure}[b]{0.15\textwidth}
        \includegraphics[width=\textwidth]{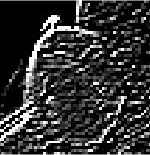}
      
    \end{subfigure}
    ~ 
    \begin{subfigure}[b]{0.15\textwidth}
        \includegraphics[width=\textwidth]{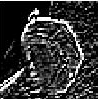}
    \end{subfigure}
  
    \caption{From left to right: a sample image from  Webcam dataset, the output of a bad filter and the reconstructed output.}\label{fig:webcam_reconstruction}
     \vspace*{-0.8cm} 
\end{figure}

\begin{table}
\caption{\label{tab2}Recognition accuracies on variety of datasets}
\centering
\begin{tabular}{ | c | c| c | c |}
\hline
Method & Mnist$\rightarrow$MnistM & Syn$\rightarrow$Dark & Photo$\rightarrow$Art \\  \hline
CNN(NA) & 54.6 & 75.0 & 85.2  \\  \hline
Filter Reconstruction &  \textbf{56.7} &  \textbf{80.0} & \textbf{ 86.7}  \\ \hline
\end{tabular}
\end{table}
 \vspace*{-0.5cm} 
 Of course in this case there is also a high level shift that can be corrected by adapting the last layer features. See Figures \ref{fig:gray_reconstruction}, \ref{fig:webcam_reconstruction} for two filter reconstruction examples from Webcam and Amazon-Gray target sets. In the case of Amazon-Gray, we have the original color image from Amazon and also each filter output which serves as a reference. More examples can be viewed in the supplementary materials.
The method also outperforms the DDC~\cite{tzeng2014deep} which is dedicated  to correct the shift at the last layer only. The same for SA where the method improvement on Amazon gray was negligible. A similar behavior can be observed for the SA applied to the first layer where we refer that to the fact that aligning all the filters of the target on one shot with the filters of the source can be seen as ignoring the bad features (filters) while in our case we try to bring up new information by reconstructing the bad filters' output.

Table \ref{tab2} shows the prediction's accuracy on the  remaining three datasets.  For Syn$\rightarrow$Dark, the method succeeds in improving the performance by 5\%.  This again proves that  the first layer filters are not generic and adapting them plays an important role in the rest of the classification process. It also indicates the direct applicability of the method in real life applications.
A similar performance can be observed w.r.t. the Photo-Art and Mmnist-MnistM adaptation problems. As a result, we can conclude that the method could improve on a variety of datasets based on different network architectures and could scale from small datasets like Office and Photo-Art up to large datasets such as the Synthetic traffic signs and Mnist.The whole procedure of filter selection and construction takes around 5 minutes on a desktop CPU Core i7 with 16G RAM. 

\paragraph{\textbf{Using few samples}}
As explained earlier, our method uses the unlabeled target samples only to estimate  the target distribution and thus the divergence of each filter. To model the distribution of a filter, we don't need a big number of samples - especially since we do not take the spatial resolution into account and consider each point in the filter map as a different sample. So, an image is composed of a number of samples equal to the size of the filter map.

\begin{figure}

\begin{minipage}[t]{.5\textwidth}

\vspace*{-0.3cm} 
\includegraphics[width=0.95\textwidth,scale=0.3]{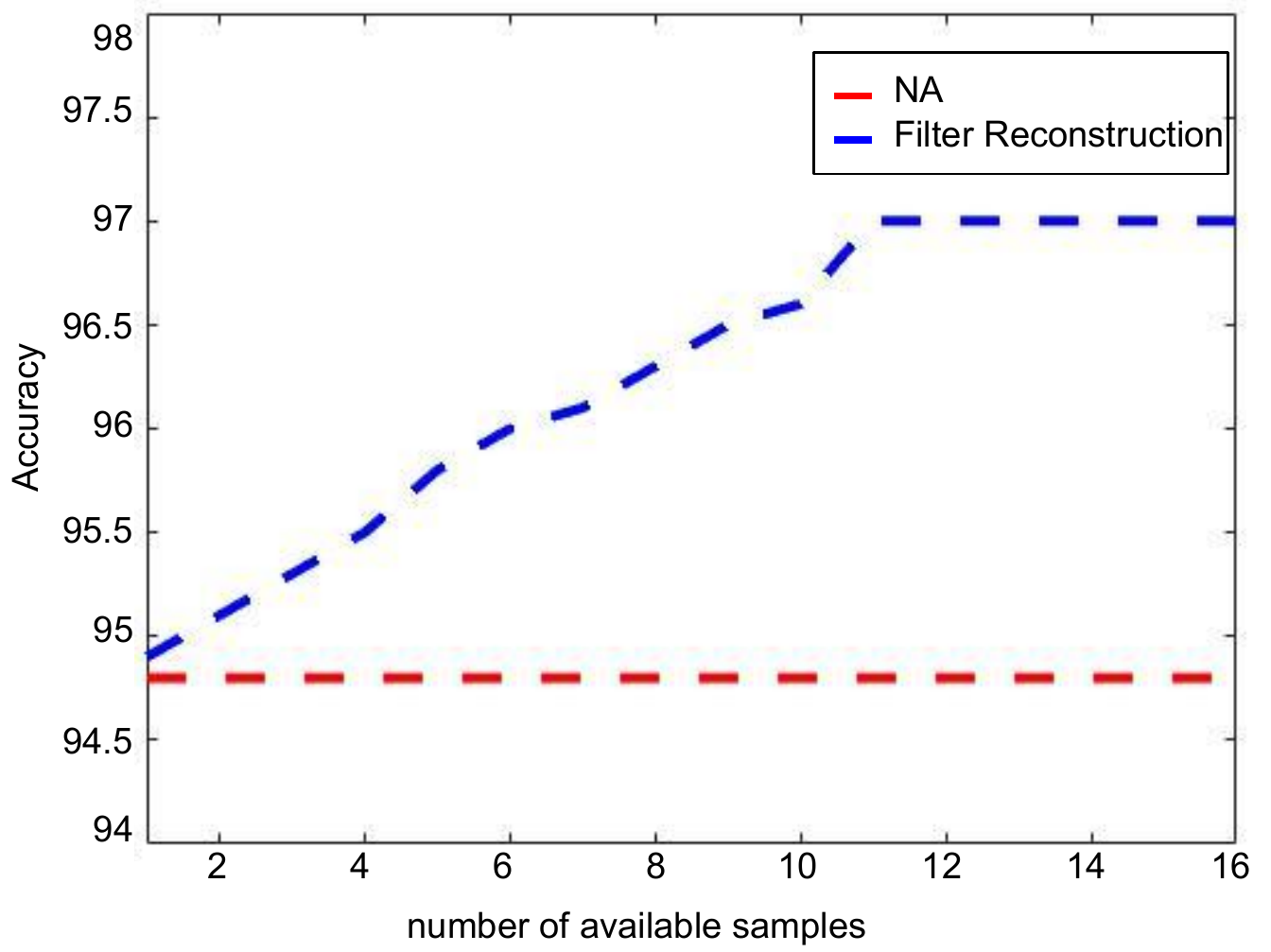}
\hspace{-1pt}

\caption{A$\rightarrow$A-G few samples adaptation}
\label{fig:fewSampels}
\vspace*{-0.5cm} 
\end{minipage}\hfill
\begin{minipage}[t]{.5\textwidth}

\vspace*{-0.3cm} 
\captionof{table}{\label{tab3} Filter reconstruction accuracy for the 2nd and 3rd layer only compared to 1st layer.}
\begin{tabular}{ | c | c| c | c |}
\hline
Method & A$\rightarrow$W& A$\rightarrow$D & A$\rightarrow$A-G \\ \hline
CNN(NA) & 60.5 & 65.8 & 94.8  \\ \hline 
1st layer &  62.0&  67.2 &  97.0  \\ \hline 
2nd layer &  61.0&  66.2 &  96.1  \\ \hline 
3rd layer &  60.8 &66.2 &  94.8  \\ \hline 
\end{tabular}
\end{minipage}
\vspace*{-0.3cm} 
\end{figure}

To examine this claim, we use sets of different numbers of images from the target dataset, here w.r.t. the Amazon-Gray dataset. We run the method 3 times for 3 different subsets 
and report the mean performance starting from 1 available target sample (see figure \ref{fig:fewSampels}). The method starts improving the performance already from a single sample and this improvement gradually increases up to 10, after which the performance saturates.

\paragraph{\textbf{Examining other layers}}
Here, we extend our test to further layers i.e.  rather than adapting the filters of the first layer, we try to reconstruct the filters of the second  and third layer instead (leaving the other layers untouched).
As reported in Table \ref{tab3}, the improvement obtained by adapting the second and third layers is less than the first layer. Especially, with Amazon-Gray, adapting the third layer features doesn't improve the recognition. This may be explained by the fact that the domain shift originates from the first layer (color) and thus can't be corrected by adapting the third layer only, which has more texture-oriented filters.

\section{Conclusion}\label{sec:Conclusion}
In this work, we aim to push the limits of unsupervised domain adaptation methods to settings where we have few samples and limited resources to adapt, both in terms of memory and time. 
To this end, we perform an extensive analysis of the output of a deep network from a domain adaptation point of view. We deduce that even though filters of the first layer seem relatively generic, domain shift issues already manifest themselves at this early stage. Therefore, we advocate that the adaptation process should start from the early  layers rather than just adapting the last layer features, as is often done in the literature. Guided by this analysis, we propose a new method that corrects the low level shift without retraining the network. The proposed method is suitable when moving a system to a new environment and can be seen as a preprocessing step that requires just a few images to be functional. This opens the door towards online adaptation, where the model is updated as new instances are becoming available. The  method is lightweight, can be applied to different tasks and is not conditioned by a specific architecture. We test it on a variety of datasets in which it systematically succeeds to improve the network performance.

\bibliographystyle{splncs}
\bibliography{egbib}
\end{document}